%% file: EMNLP2021.tex
\pdfoutput=1
\documentclass[11pt]{article}

\usepackage{emnlp2021}

\usepackage{times}
\usepackage{latexsym}

\usepackage[T1]{fontenc}
\usepackage[utf8]{inputenc}

\usepackage{microtype}

\usepackage{graphicx} 
\usepackage{float}
\usepackage{subfigure}
\usepackage{amsmath}
\usepackage{amssymb}
\usepackage{booktabs}
\usepackage{color}
\usepackage{makecell}

\usepackage{xurl}

\title{Abstract, Rationale, Stance: \\ A Joint Model for Scientific Claim Verification}

\author{Zhiwei Zhang$^{1,2}$,
        Jiyi Li$^{2}$\footnote{Corresponding author},
        Fumiyo Fukumoto$^{2}$ and Yanming Ye$^{1}$ \\
        Hangzhou Dianzi University, Hangzhou, China$^{1}$ \\
        University of Yamanashi, Kofu, Japan$^{2}$\\
        \texttt{hduxiaozhi97@gmail.com, \{jyli,fukumoto\}@yamanashi.ac.jp} \\ \texttt{yeym@hdu.edu.cn}}

\begin{document}
\maketitle

\begin{abstract}
\input{contents/abstract}
\end{abstract}

\section{Introduction}
\input{contents/introduction}

\section{Our Approach}
\input{contents/approach}

\section{Experiments}
\input{contents/experiments}

\section{Conclusion}
\input{contents/conclusion}

\section*{Acknowledgments}
This work was partially supported by KDDI Foundation Research Grant Program.

\clearpage
\bibliography{custom}
\bibliographystyle{acl_natbib}

\end{document}

%% file: contents/abstract.tex
Scientific claim verification can help the researchers to easily find the target scientific papers with the sentence evidence from a large corpus for the given claim.~Some existing works propose pipeline models on the three tasks of abstract retrieval, rationale selection and stance prediction.~Such works have the problems of error propagation among the modules in the pipeline and lack of sharing valuable information among modules. We thus propose an approach, named as \textsc{ARSJoint}, that jointly learns the modules for the three tasks with a machine reading comprehension framework by including claim information. 
In addition, we enhance the information exchanges and constraints among tasks by proposing a regularization term between the sentence attention scores of abstract retrieval and the estimated outputs of rational selection. The experimental results on the benchmark dataset \textsc{SciFact} show that our approach outperforms the existing works.

%% file: contents/introduction.tex
\begin{figure}[!t]
\centering 
\includegraphics[width=0.45\textwidth]{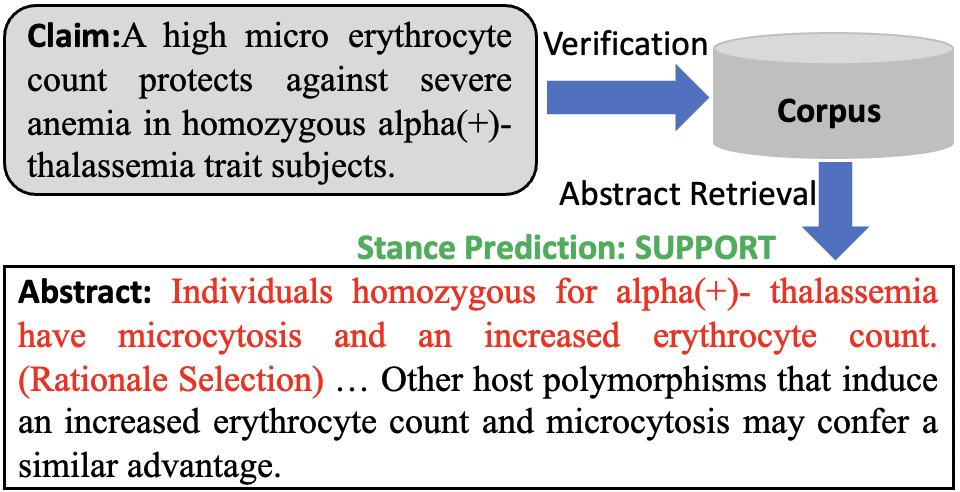}
\caption{An example of scientific claim verification. }
\label{fig:example}
\end{figure}

A system of scientific claim verification can help the researchers to easily find the target scientific papers with the sentence evidence from a large corpus for the given claim. To address this issue, \citet{wadden2020fact} introduced scientific claim verification which consists of three tasks. As illustrated in Figure \ref{fig:example}, for a given claim, the system finds the abstracts which are related to the claim from a scholarly document corpus (abstract retrieval task); it selects the sentences which are the evidences in the abstract related to the claim (rationale selection task); it also classifies whether the abstract/sentences support or refute the claims (stance prediction task). \citet{wadden2020fact} also provided a dataset called \textsc{SciFact}. 

Most of the existing works of \textit{general} claim verification are based on pipeline models~\citep{soleimani2020bert,alonso2019team,liu2020fine,zhou2019gear,nie2019combining,lee2020misinformation}; some works utilize joint optimization strategies~\citep{lu2020gcan,yin2018twowingos,hidey-etal-2020-deseption}. These models attempted to jointly optimize the rationale selection and stance prediction, but did not directly link the two modules \citep{li2020paragraph}.~In the case of the \textit{scientific} claim verification, \citet{wadden2020fact} proposed a baseline model \textsc{VeriSci} based on a pipeline of three components for the three tasks. 
\citet{pradeep2020scientific} proposed a pipeline model called \textsc{VerT5erini} which utilized the pre-trained language model T5 ~\citep{raffel2019exploring} and adapted a pre-trained sequence-to-sequence model. \citet{li2020paragraph} jointly trained two tasks of rationale selection and stance prediction, and had a pipeline on abstract retrieval task and the joint module.

Above existing works on scientific claim verification are fully or partially pipeline solutions. One problem of these works is the error propagation among the modules in the pipeline. Another problem is that the module in the pipeline trained independently cannot share and leverage the valuable information among each other. Therefore, we propose an approach, named as \textsc{ARSJoint}, which jointly learns the three modules for the three tasks. It has a Machine Reading Comprehension (MRC) framework which uses the claim content as the query to learn additional information. In addition, we assume that the abstract retrieval module should have good interpretability and tend to assign high sentence-level attention scores to the evidence sentences that influence the retrieval results; it is consistent with the goal of the rationale selection module. We thus enhance the information exchanges and constraints among tasks by proposing a regularization term based on a symmetric divergence to bridge these two modules. 

The experimental results on the benchmark dataset \textsc{SciFact} show that the proposed approach has better performance than the existing works. The main contributions of this paper can be summarized as follows. (1). We propose a scientific claim verification approach which jointly trains on the three tasks in a MRC framework. (2). We propose a regularization based on the divergence between the sentence attention of abstract retrieval and the outputs of rational selection.

%% file: contents/approach.tex
\subsection{Notation and Definitions}

We denote the query claim as $q$ and an abstract of a scientific paper as $a \in \mathcal{A}$. We denote the set of sentences in abstract $a$ as $\mathcal{S}=\{s_{i}\}^l_{i=1}$ and the word sequence of $s_i$ is $[s_{i1},...,s_{in_i}]$. The title of the paper $t \in \mathcal{T}$ is used as auxiliary information, the word sequence of $t$ is $[t_1,...,t_{n_t}]$. Here, $\mathcal{S}$, $s_i$ and $t$ are for $a$ in default and we omit the subscripts `$a$' in the notations. The purpose of the \textit{abstract retrieval task} is to detect the set of related abstracts to $q$; it assigns relevance labels $y^b \in \{0,1\}$ to a candidate abstract $a$. The \textit{rationale selection} task is to detect the decisive rationale sentences $S^r \subseteq S$ of $a$ relevant to the claim $q$; it assigns evidence labels $y^r_i \in \{0,1\}$ to each sentence $s_i\in \mathcal{S}$. The \textit{stance prediction} task classifies $a$ into stance labels $y^e$ which are in \{\text{SUPPORTS}=0, \text{REFUTES}=1, \text{NOINFO}=2\}. The sentences in $a$ have the same stance label value. 

\begin{figure*}[]
\centering 
\includegraphics[width=0.8\textwidth]{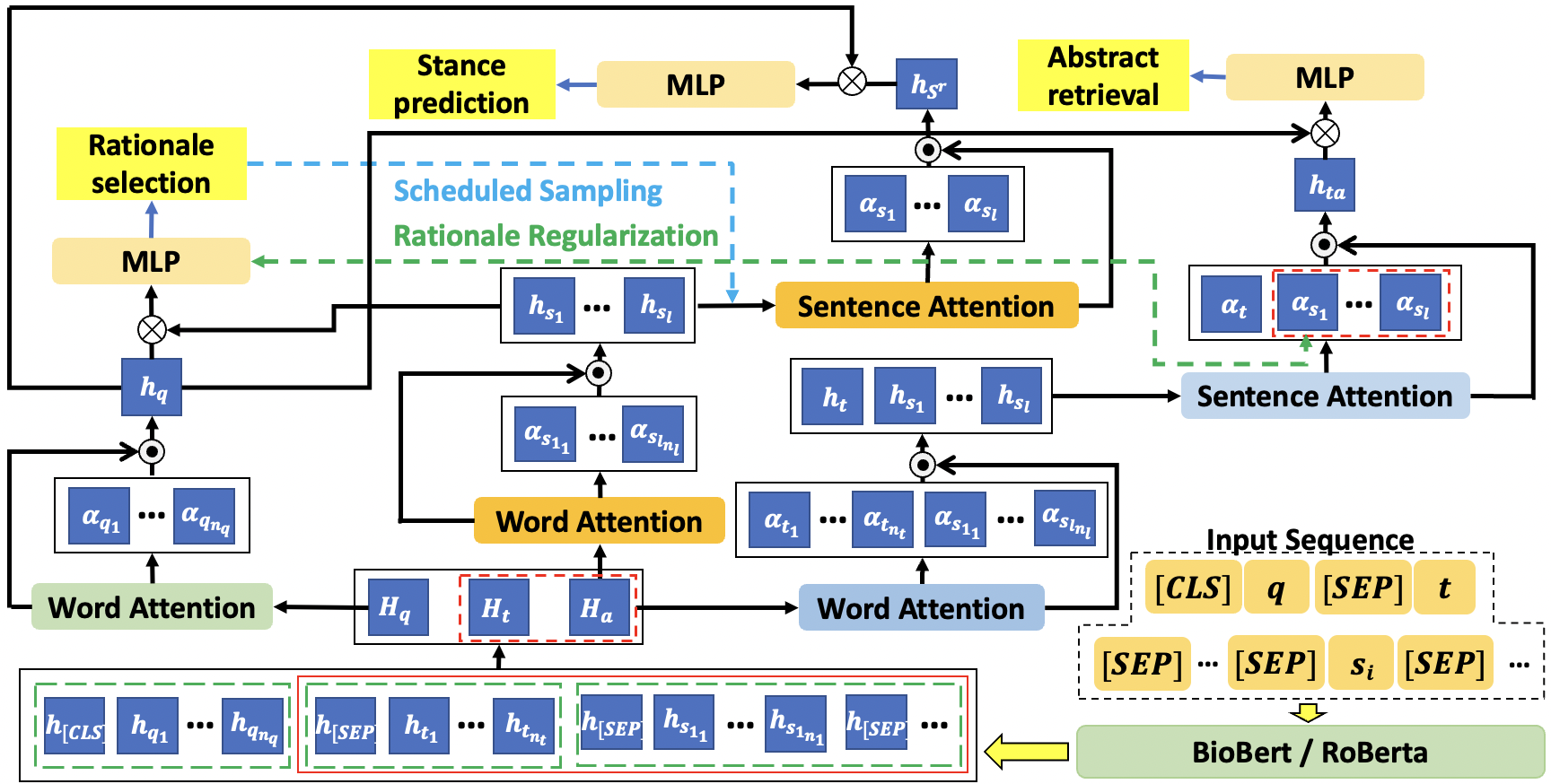}
\caption{Framework of our \textsc{ARSJoint} model which jointly learns three modules and has rationale regularization. }
\label{fig:model}
\end{figure*}

\subsection{Pre-processing}

As there are a huge amount of papers in the corpus, applying all of them to the proposed model is time-consuming. Therefore, similar to the existing works on this topic \citep{wadden2020fact, pradeep2020scientific,li2020paragraph}, we also utilize a lightweight method to first roughly select a set of candidate papers. We used the BioSentVec ~\citep{chen2019biosentvec, pagliardini2017unsupervised} to obtain the embeddings of the claim or a scientific paper based on its title and abstract, and compute the cosine similarity between the claim and the paper. The papers with top-$k$ similarities are used as the candidates.

\subsection{Joint Abstract, Rationale, Stance Model}
The input sequence of our model is defined as $seq=[[\text{CLS}]q[\text{SEP}]t\cdots[\text{SEP}]s_i [\text{SEP}]\cdots]$,~which is obtained by concatenating the claim $q$, title $t$ and abstract $a$.~We compute the list of word representations $\mathbf{H}_{seq}$ of the input sequence by a pre-trained language model (e.g., BioBERT~\citep{lee2020biobert}). We obtain the word representations of the claim $\mathbf{H}_{q} = [\mathbf{h}_{q_1}, \cdots, \mathbf{h}_{q_{n_q}}]$, the title $\mathbf{H}_{t} = [\mathbf{h}_{t_1}, \cdots, \mathbf{h}_{t_{n_t}}]$, each sentence $\mathbf{H}_{s_i} = [\mathbf{h}_{s_{i1}}, \cdots, \mathbf{h}_{s_{in_i}}]$, and the abstract $\mathbf{H}_{S} = \mathbf{H}_{a} = [\cdots, \mathbf{H}_{s_{i}}, \cdots]$ from $\mathbf{H}_{seq}$ and use them in our \textsc{ARSJoint} model. Figure \ref{fig:model} shows the framework of our model with three modules for the three tasks. 

In all three modules, we use attention layer (denoted as $g(\cdot)$) on word (sentence) representations to compute a sentence (document) representation. A document can be a claim, title, abstract, or their combinations. The computation is as follows (refer to \citep{li2020paragraph}), where the * in $\mathbf{H}_{*}$ represents any type of sentence (claim $q$, title $t$ or a sentence $s$ in an abstract), the $\star$ in  $\mathbf{H}_{\star}$ represents any type of document, $\mathbf{W}$ and $\mathbf{b}$ are trainable parameters. 
\begin{equation}
\small
    \begin{gathered}
    g(\mathbf{H}_{*})=\sum\nolimits_i{\mathbf{u}_{*_i} \boldsymbol{\alpha}_{*_i}}, \ 
    \boldsymbol{\alpha}_{*_i} = \frac{\exp(\mathbf{W}_{w_2}\mathbf{u}_{*_i} + \mathbf{b}_{w_2})}{\sum\nolimits_{j}{\exp(\mathbf{W}_{w_2}}\mathbf{u}_{*_j} + \mathbf{b}_{w_2})}, \\ \mathbf{u}_{*_j} = \tanh(\mathbf{W}_{w_1} \mathbf{h}{*_j} + \mathbf{b}_{w_1}) \text{ for word-level attention, } \\ 
    g(\mathbf{H}_{\star})=\sum\nolimits_i{\mathbf{U}_{\star_i} \boldsymbol{\alpha}_{\star_i}}, \ 
    \boldsymbol{\alpha}_{\star_i} = \frac{\exp(\mathbf{W}_{c_2}\mathbf{U}_{\star_i} + \mathbf{b}_{c_2})}{\sum\nolimits_{j}\exp(\mathbf{W}_{c_2}\mathbf{U}_{\star_j} + \mathbf{b}_{c_2})}, \\ 
    \mathbf{U}_{\star_j} = \tanh(\mathbf{W}_{c_1} \mathbf{H}{\star_j} + \mathbf{b}_{c_1})  \text{ for sentence-level attention}.\\
    \end{gathered}
\label{eq:attentin}
\end{equation}

\noindent
\textbf{Abstract Retrieval:} 
In this task, a title can be regarded as an auxiliary sentence that may contain the information related to the claim for the abstract, we thus use the title with the sentences in the abstract together. 
We build a document $ta=[t,a]$ and concatenate the word representations of $t$ and $a$ into $\mathbf{H}_{ta} = [\mathbf{H}_{t}, \mathbf{H}_{a}]$ as the input to this module. We use a hierarchical attention network ($\text{HAN}$) ~\citep{yang2016hierarchical} to compute document representations $\mathbf{h}_{ta}\in \mathbb{R}^d$, $\mathbf{h}_{ta} = \text{HAN}(\mathbf{H}_{ta})$. $\text{HAN}$ is proper for document classification by considering the hierarchical document structure (a document has sentences, a sentence has words). We also compute the sentence representation of claim $\mathbf{h}_{q}\in \mathbb{R}^d$ with a word-level attention layer (denoted as $g(\cdot)$), $\mathbf{h}_{q} = g(\mathbf{H}_{q})$. To compute the relevance between $\mathbf{h}_{ta}$ and $\mathbf{h}_{q}$, we use a Hadamard product on them and a Multi-Layer Perception ($\text{MLP}$, denoted as $f(\cdot)$) with $\text{Softmax}$ (denoted as $\sigma(\cdot)$); the outputs are the probabilities that whether the abstract is relevant to the claim, $[p^b_0, p^b_1] =  \sigma(f(\mathbf{h}_q \circ \mathbf{h}_{a}))$. A cross entropy loss $\mathcal{L}_{ret}$ is used for training. 

\noindent
\textbf{Rationale Selection:} This task focuses on judging whether a sentence in the abstract is a rationale one or not. For the multiple sentences in the abstract, they have same title information but have different rationale labels. Therefore, when judging each sentence in the abstract, using the title may not positively influence the performance. We thus use the word representation $\mathbf{H}_{a}$ of the abstract as input. 
We compute the sentence representation $\mathbf{h}_{s_i}$ by a word-level attention layer, and use a MLP with Softmax to estimate the probability $p^{r}_{i1}$ and $p^{r}_{i0}$ that whether $s_i$ is the evidence of the abstract or not. The cross entropy loss is $\mathcal{L}_{rat}$. 

\noindent
\textbf{Stance Prediction:} The module first computes the sentence representation $\mathbf{h}_{s_i}$ in a same way with that of rationale selection. After that, it only selects the sentences $S^r$ with the true evidence label $\hat{y}^r_i=1$ or the estimated evidence probability $p^r_{i1}>p^r_{i0}$; whether using the true label or the estimated label is decided by a scheduled sampling which will be introduced later. We then compute the estimated stance labels based on a sentence-level attention layer and a MLP with Softmax, $\mathbf{h}_{S^r} = g(\mathbf{H}_{S^r})$ and $[p^e_0, p^e_1, p^e_2] = \sigma(f(\mathbf{h}_q \circ\mathbf{h}_{S^r}))$, where $S^r=\{s_i\in S|\hat{y}^r_i=1 \text{ or } p^r_{i1}>p^r_{i0}\}$. The cross entropy loss is $\mathcal{L}_{sta}$.  

\noindent
\textbf{Scheduled Sampling:} Since rationale sentences $S^r$ are used in stance prediction, the error of the rationale selection module will be propagated to the stance prediction module. To alleviate this problem, following~\citep{li2020paragraph}, we also use a scheduled sampling method \citep{bengio2015scheduled}, which is to feed the sentences with true evidence label $\hat{y}^r_i=1$ to the stance prediction module at the beginning, and then gradually increase the proportion of the sentences with the estimated evidence probability $p^r_{i1}>p^r_{i0}$, until eventually all sentences in $S^r$ are based on the estimated evidences. We set the sampling probability of using the estimated evidences as $p_{sample} = \sin(\frac{\pi}{2}\times\frac{current\_epoch-1}{total\_epoch-1})$.

\begin{table}[!t]
\centering
\footnotesize
\begin{tabular}{c|ccc|c}
\hline
\multicolumn{1}{l|}{} & SUPPORT & NOINFO & REFUTES & ALL  \\ \hline
Train                 & 332 / 370     & 304 / 220           & 173 / 194        & 809  \\
Dev.                   & 124 / 138     & 112 / 114           & 64 / 71         & 300  \\
\hline
ALL                   & 456 / 508     & 416 / 444           & 237 / 265        & 1109
\\ \hline
\end{tabular}
\caption{Statistics of \textsc{SciFact} dataset. The numbers are "number of claims / number of relevant abstracts".}
\label{tab:dataset}
\end{table}

\noindent
\textbf{Rationale Regularization (RR):} The attention scores have been used for interpretability in NLP tasks \citep{serrano-smith-2019-attention,wiegreffe-pinter-2019-attention,sun-lu-2020-understanding}. 
We assume that the abstract retrieval module should have good interpretability and tend to assign high sentence-level attention scores to the evidence sentences that influence the retrieval results; it is consistent with the goal of the rationale selection module. We thus enhance the information exchanges and constraints among tasks by proposing a regularization term based on a symmetric divergence on the sentence attention scores $\boldsymbol{\alpha}$ of abstract retrieval and the estimated outputs $\mathbf{y}^r$ of the rational selection to bridge these two modules. The detailed formula is as follows, where $\mathbf{p}$ and $\mathbf{q}$ are $\boldsymbol{\alpha}$ or $\mathbf{y}^r$. 
\begin{equation}
\small
    \begin{gathered}
    \mathcal{D}(\mathbf{p}||\mathbf{q}) = -\sum_{i=1}^l\left(\mathbf{p}_i\log(\mathbf{q}_i)+(1-\mathbf{p}_i)\log(1-\mathbf{q}_i)\right), \\
    \mathcal{L}_{RR} = \mathcal{D}(\boldsymbol{\alpha}||\mathbf{y}^r) + \mathcal{D}(\mathbf{y}^r||\boldsymbol{\alpha}). 
    \end{gathered}
\label{eq:loss}
\end{equation}

\noindent
\textbf{Joint Training:} 
We jointly train our model on abstract retrieval, rationale selection and stance prediction. The joint loss with our RR is as follows, $\mathcal{L} = \lambda_1\mathcal{L}_{ret} + \lambda_2\mathcal{L}_{rat} + \lambda_3\mathcal{L}_{sta} +\gamma \mathcal{L}_{RR}$, where $\lambda_1$, $\lambda_2$, $\lambda_3$ and $\gamma$ are hyperparameters.

%% file: contents/experiments.tex
\subsection{Experimental Settings}

\noindent
\textbf{Dataset:} We utilize the benchmark dataset \textsc{SciFact}\footnote{\label{fn:SciFact}\url{https://github.com/allenai/scifact}}. It consists of 5,183 scientific papers with titles and abstracts and 1,109 claims in the training and development sets. Table~\ref{tab:dataset} presents the statistics of the dataset. 

\noindent
\textbf{Experimental Settings:}
For our \textsc{ARSJoint} model, we use Optuna~\citep{akiba2019optuna} to tune the hyperparameters $\lambda_{1}$, $\lambda_{2}$, $\lambda_{3}$ and $\gamma$ of the loss $\mathcal{L}$ on 20\% of the training set and based on the performance on another 20\% training set. We choose the optimal hyperparameters by the average F1-score on abstract-level and sentence-level evaluations. The search ranges of these four hyperparameters are set to [0.1, 12], and the number of search trials is set to 100. Table~\ref{tab:Hyper-parameters-loss} lists the selected weight hyperparameters of our model. The other hyperparameters such as learning rate in the model refer to the ones used in exiting work~\citep{li2020paragraph} to make a fair comparison. These hyperparameters are listed in Table~\ref{tab:Hyper-parameters}. 

We implement our \textsc{ARSJoint} model\footnote{\label{fn:ARSJoint}Our source code is available at:  \url{https://github.com/ZhiweiZhang97/ARSJointModel}} in PyTorch. Since the length of the input sequence $seq$ is often greater than the maximum input length of a BERT-based model, we perform a tail-truncation operation on each sentence of $seq$ that exceeds the maximum input length. For the pre-trained language model, we verify our approach by respectively using RoBERTa-large~\citep{liu2019roberta} and BioBERT-large~\citep{lee2020biobert} trained on a biomedical corpus. We fine-tune RoBERTa-large and BioBERT-large on the \textsc{SciFact} dataset. 
In addition, the MLP in our model has two layers. 

We compare our \textsc{ARSJoint} approach with  Paragraph-Joint~\citep{li2020paragraph}, \textsc{VeriSci}$^{\ref{fn:SciFact}}$~\citep{wadden2020fact} and \textsc{VerT5erini}~\citep{pradeep2020scientific}. We use the publicly available code$^{\ref{fn:ARSJoint}}$ of them. The "Paragraph-Joint Pre-training" model is pre-trained on the FEVER dataset~\citep{thorne2018fever} and then fine-tune on the \textsc{SciFact} dataset. The "Paragraph-Joint \textsc{SciFact}-only" is not pre-trained on other datasets. 

\begin{table}[!t]
    \setlength{\tabcolsep}{1.8mm}
    \footnotesize
    \centering
    \begin{tabular}{c|cccc}
        \hline
         Model &	$\lambda_1$ &	$\lambda_2$ &	$\lambda_3$ & 	$\gamma$ \\ \hline
        \textsc{ARSJoint} w/o RR (RoBERTa) &	2.7 &	11.7 &	2.2 &	- \\
        \textsc{ARSJoint} (RoBERTa) &	0.9 &	11.1 &	2.6 &	2.2 \\
        \textsc{ARSJoint} w/o RR (BioBERT) &	0.1 &	10.8 &	4.7 &	- \\
        \textsc{ARSJoint} (BioBERT) &	0.2 &	12.0 &	1.1 &	1.9 \\
        \hline
    \end{tabular}
    \caption{Hyperparameters selected by Optuna for different variants of our model. The "w/o RR" means the model does not utilize rationale regularization.}
    \label{tab:Hyper-parameters-loss}
\end{table}

\begin{table}[!t]
    \setlength{\tabcolsep}{1.75mm}
    \footnotesize
    \centering
    \begin{tabular}{c|c|c|c|c|c}
    \hline
    Name & Value & Name        & Value & Name        & Value   \\ \hline
    $k_{tra}$      & 12    & $lr_{1}$ & $1\times10^{-5}$ & Batch size  & 1 \\
    $k_{ret}$  & 30    & $lr_{2}$      & $5\times10^{-6}$ & Dropout            & 0 \\\hline
    \end{tabular}
    \caption{Hyperparameter settings following the existing work. $k_{tra}$ and $k_{ret}$ are the number of candidate abstracts for each claim in the training and testing stages. $lr_{1}$ and $lr_{2}$ are the learning rates of the BERT-based model and other modules of the proposed model.}
    \label{tab:Hyper-parameters}
\end{table}

\noindent
\textbf{Evaluation:} We evaluate the methods by using the abstract-level and sentence-level evaluation criteria given in \textsc{SciFact}$^{\ref{fn:SciFact}}$. 
\textit{Abstract-level evaluation:} It evaluates the performance of a model on detecting the abstracts which support or refute the claims. For the "Label-Only" evaluation, given a claim $q$, the classification result of an abstract $a$ is correct if the estimated relevance label $\hat{y}^b$ is correct and the estimated stance label $\hat{y}^e$ is correct. For the "Label+Rationale" evaluation, the abstract is correctly rationalized, in addition, if the estimated rationale sentences contain a gold rationale. 
\textit{Sentence-level evaluation:} It evaluates the performance of a model on detecting rationale sentences. For the "Selection-Only" evaluation, an estimated rationale sentence $s_i$ of an abstract $a$ is correctly selected if the estimated rationale label $\hat{y}^r_i$ is correct and the estimated stance label $\hat{y}^e$ is not "NOINFO". Especially, if consecutive multiple sentences are gold rationales, then all these sentences should be estimated as rationales. For the "Selection+Label", the estimated rationale sentences are correctly labeled, in addition, if the estimated stance label $\hat{y}^e$ of this abstract is correct. The evaluation metrics F1-score (F1), Precision (P), and Recall (R) are used. We train the model using all training data, and since \citet{wadden2020fact} does not publish the labels on the test set, we evaluate the approaches on the development set following \citep{li2020paragraph}.

\setlength{\tabcolsep}{1.9mm}{
\begin{table*}[!t]
\centering
\footnotesize
\begin{tabular}{l|c c c|c c c|c c c|c c c}
\hline
& \multicolumn{6}{c|}{{Sentence-level}}  & \multicolumn{6}{c}{{Abstract-level}} \\ 
& \multicolumn{3}{c|}{{Selection-Only}} & \multicolumn{3}{c|}{{Selection+Label}}             & \multicolumn{3}{c|}{{Label-Only}} & \multicolumn{3}{c}{{Label+Rationale}} \\  \cline{2-13} 
{{Models}} & {P} & {R} & {F1} & {P} & {R} & {F1} & {P} & {R} & {F1} & {P} & {R} & {F1} \\ \hline
{\textsc{VeriSci}}                        &  54.3 &  43.4 &  48.3 &  48.5 &  38.8 &  43.1 &  56.4 &  48.3 &  52.1 &  54.2 &  46.4 &  50.0 \\ 
{Paragraph-Joint \textsc{SciFact}-only} &  69.3 &  50.0 &  58.1 &  59.8 &  43.2 &  50.2 &  69.9 &  52.1 &  59.7 &  64.7 &  48.3 &  55.3 \\ 
{Paragraph-Joint Pre-training}                &  74.2 &  57.4 &  64.7 &  63.3 &  48.9 &  55.2 &  71.4 &  59.8 &  65.1 & {65.7} &  55.0 &  59.9 \\  
{\textsc{VerT5erini}   (BM25)}           &  67.7 &  53.8 &  60.0 &  63.9 &  50.8 &  56.6 &  70.9 &  61.7 &  66.0 &  67.0 &  58.4 &  62.4 \\ 
{\textsc{VerT5erini}   (T5)}             &  64.8 &  57.4 &  60.9 &  60.8 &  \textbf{53.8} &  57.1 &  65.1 &  \textbf{65.1} &  65.1 &  61.7 &  \textbf{61.7} &  61.7 \\ \hline \hline

{\textsc{ARSJoint} w/o RR (RoBERTa)}     & 70.9 & 56.6 & 62.9 & 56.8 & 45.4 & 50.5 & 66.1 & 56.0 & 60.6 & 61.0 & 51.7 & 56.0 \\ 
{\textsc{ARSJoint} (RoBERTa)}     & 67.9 & 57.1 & 62.0 & 55.5 & 46.7 & 50.7 & 64.5 & 57.4 & 60.8 & 59.1 & 52.6 & 55.7 \\ 
{\textsc{ARSJoint} w/o RR (BioBERT)}     & 75.4 & 57.7 & 65.3 & 63.6 & 48.6 & 55.1 & 72.7 & 57.4 & 64.2 & 67.9 & 53.6 & 59.9 \\ 
{\textsc{ARSJoint} (BioBERT)}     & \textbf{76.2} & \textbf{58.5} & \textbf{66.2} & \textbf{66.5} & 51.1 & \textbf{57.8} & \textbf{75.3} & 59.8 & \textbf{66.7} & \textbf{70.5} & 56.0 & \textbf{62.4} \\ 
\hline
\end{tabular}
\caption{Main experimental results. }
\label{tab:result}
\end{table*}
}

\linespread{1.1}
\setlength{\tabcolsep}{1.9mm}{
\begin{table*}[]
\centering
\footnotesize
\begin{tabular}{m{12.7cm}|c|c|c}
\hline
\textbf{Claim:} Ly6C hi monocytes have a lower inflammatory capacity than Ly6C lo monocytes. & $\alpha_i$ & $\hat{y}^r_i$ & $y^r_i$ \\ \hline\hline
Blood monocytes are well-characterized precursors for macrophages and dendritic cells. &0.0745 &0 &0 \\ \hline
\multicolumn{4}{c}{......} \\
\hline
\textcolor[rgb]{1,0,0}{Under inflammatory conditions elicited either by acute infection with Listeria monocytogenes or chronic 1,0,0 infection with Leishmania major, there was a significant increase in immature Ly-6C(high) monocytes, resembling the inflammatory left shift of granulocytes.} & 0.0936& 1 &1\\ \hline
\textcolor[rgb]{1,0,0}{In addition, acute peritoneal inflammation recruited preferentially Ly-6C(med-high) monocytes.}& 0.1613&1 &1 \\ \hline
Taken together, these data identify distinct subpopulations of mouse blood monocytes that differ in maturation stage and capacity to become recruited to inflammatory sites. &0.0745 &0 &0 \\
\hline
\end{tabular}
\caption{Result example of Rationale Regularization. Given a claim, it lists the sentences from an abstract. $\alpha_i$ is sentence attention score in the abstract retrieval task; $\hat{y}^r_i$ is estimated rationale label; $y^r_i$ is true rationale label. } 
\label{tab:RR}
\end{table*}
}

\subsection{Experimental Results}
Table~\ref{tab:result} shows the main experimental results. First, the proposed method \textsc{ARSJoint} (BioBERT) outperforms the existing works with fully or partially pipelines.~\textsc{VeriSci} and \textsc{VerT5erini} are pipeline models and Paragraph-Joint is a partially pipeline model with a joint model on two tasks. It shows that the proposed model which jointly learns the three tasks is effective to improve the performance. 

Second, when using the same pre-trained model RoBERTa-large, comparing our method and the paragraph-joint model, \textsc{ARSJoint} (RoBERTa) and \textsc{ARSJoint} w/o RR (RoBERTa) have better performance than "Paragraph-Joint SciFact Only", especially on Recall. It shows that jointly learning with the abstract retrieval task can improve performance. For the Paragraph-Joint method, "Paragraph-Joint Pre-training" with pre-training on another FEVER dataset has much better performance than "Paragraph-Joint \textsc{SciFact}-only" without pre-training on other datasets. Similarly, we replace RoBERTa-large with BioBERT-large which contains biological knowledge;  \textsc{ARSJoint} (BioBERT) achieves better performance over "Paragraph-Joint Pre-training". 

Third, as an ablation study of the proposed RR, in the case of using BioBERT-large, there is a significant difference between the model with and without RR. Although only a small difference in the case of using RoBERTa-large, there is still an improvement on Recall. This indicates that rationale regularization can effectively improve the performance of the model. Table~\ref{tab:RR} shows an example of the results with RR. In this example, it lists a claim and the sentences from an abstract. The attention scores of the sentences in the abstract retrieval task are consistent with the true rationale labels (as well as the estimated rationale labels). The abstract retrieval module thus has good interpretability.

%% file: contents/conclusion.tex
In this paper, we propose a joint model named as \textsc{ARSJoint} on three tasks of abstract retrieval, rationale selection and stance prediction for scientific claim verification in a MRC framework by including claim. We also propose a regularization based on the divergence between the sentence attention of the abstract retrieval task and the outputs of the rational selection task. The experimental results illustrate that our method achieves better results on the benchmark dataset \textsc{SciFact}. In future work, we will try to pre-train the model on other general claim verification datasets such as FEVER \cite{thorne2018fever} to improve the performance. 